\documentclass{article}

\usepackage[final]{neurips_2021}

\usepackage[utf8]{inputenc} % allow utf-8 input
\usepackage[T1]{fontenc}    % use 8-bit T1 fonts
\usepackage{hyperref}       % hyperlinks
\usepackage{url}            % simple URL typesetting
\usepackage{booktabs}       % professional-quality tables
\usepackage{amsfonts}       % blackboard math symbols
\usepackage{nicefrac}       % compact symbols for 1/2, etc.
\usepackage{microtype}      % microtypography
\usepackage{xcolor}         % colors
\usepackage{graphicx}
\usepackage{amsfonts, amsmath}
\usepackage{wrapfig}

\DeclareMathOperator{\sign}{sgn}

\title{Capturing implicit hierarchical structure in 3D biomedical images with self-supervised hyperbolic representations}

\author{%
  Joy Hsu\thanks{These authors contributed equally.} \\
  Department of Computer Science\\
  Stanford University\\
  \texttt{joycj@stanford.edu} \\
  \And
  Jeffrey Gu$^*$ \\
  ICME \\
  Stanford University\\
  \texttt{jeffgu@stanford.edu} \\
  \And
  Gong Her Wu \\
  Department of Bioengineering \\
  Stanford University\\
  \texttt{wukon@stanford.edu} \\
  \And
  Wah Chiu \\
  Department of Bioengineering \\
  Stanford University\\
  \texttt{wahc@stanford.edu} \\
  \And
  Serena Yeung \\
  Department of Biomedical Data Science \\
  Stanford University\\
  \texttt{syyeung@stanford.edu} \\
}

\begin{document}

\maketitle

\begin{abstract}
We consider the task of representation learning for unsupervised segmentation of 3D voxel-grid biomedical images. We show that models that capture \textit{implicit} hierarchical relationships between subvolumes are better suited for this task. To that end, we consider encoder-decoder architectures with a hyperbolic latent space, to explicitly capture hierarchical relationships present in subvolumes of the data. We propose utilizing a 3D hyperbolic variational autoencoder with a novel gyroplane convolutional layer to map from the embedding space back to 3D images. To capture these relationships, we introduce an essential self-supervised loss---in addition to the standard VAE loss---which infers approximate hierarchies and encourages implicitly related subvolumes to be mapped closer in the embedding space. We present experiments on both synthetic data and biomedical data to validate our hypothesis. 
\end{abstract}

\section{Introduction}
Advances in biomedical imaging techniques such as cryogenic electron tomography (cryo-ET) and magnetic resonance imaging (MRI) have resulted in an ever-increasing amount of 3D biomedical image data. In these data domains, a growing body of work shows that, when provided with labels, machine learning models achieve good performance on many tasks \citep{cciccek20163d, milletari2017hough, dou20173d, falk2019u}. However, these labels, especially for segmentation, are very costly as they often have to be provided by experts in the appropriate field. Consequently, supervised learning and even semi-supervised learning remain limited in this setting as (1) tasks and domains are very diverse, making it intractable for experts to provide labelled data for all problems; and (2) experts can only label objects they already know, restricting the potential for scientific discovery using machine learning methods. In this work, we tackle the task of unsupervised segmentation in 3D biomedical image data.
%In this work, we tackle the task of unsupervised segmentation when provided with 3D voxel-grid data from the biomedical domain.

Our key insight is that 3D biomedical images have inherent hierarchical structure. For example, in the cryo-ET domain, an image of a cell has a hierarchy that at the highest level comprises the entire cell; at a finer level comprises organelles such as the mitochondria and nucleus; and at an even finer level comprises sub-structures such as the nucleolus of a nucleus or protein machineries within organelles. Such types of hierarchies are present in many types of biomedical images (e.g., nested anatomical structures within MRI images). We hypothesize that in the unsupervised setting, models that are able to encode this internal hierarchical structure will provide better data representations for downstream tasks. 
%As such, we posit that, in the absence of labels, models that are able to encode these hierarchies will provide better data representations for downstream tasks. 
To that end, we propose learning representations based on embedding subvolumes of 3D images in hyperbolic space.

%In contrast to traditional (Euclidean) embeddings, hyper- bolic embeddings better preserve hierarchical relationships present in the data. To provide some intuition, when em- bedding a tree in hyperbolic space, the distance between nodes at increasing levels of depth grows exponentially large (Sarkar, 2011). A recent line of work utilizes hyper- bolic representations to represent hierarchical data across domains ranging from natural language word taxonomies (Nickel & Kiela, 2017; 2018) and graphs (Nickel & Kiela, 2017; Mathieu et al., 2019; Ovinnikov, 2019; Chami et al., 2019), to image classification (Mathieu et al., 2019).

In contrast to traditional Euclidean embeddings, hyperbolic embeddings better preserve hierarchical relationships present in the data. Hyperbolic representations have been proposed as a continuous way to represent hierarchical data, due to their ability to embed trees with arbitrarily low error \citep{sarkar2011low}. A recent line of work utilizes hyperbolic representations to model hierarchical data across domains ranging from natural language word taxonomies \citep{nickel2017poincare, nickel2018learning} and graphs \citep{nickel2017poincare, mathieu2019poincare, ovinnikov2019poincar, chami2019hgcn}, to image classification \citep{mathieu2019poincare}. In these settings, the objects, and in most cases their relationships, are explicitly encoded in the data. However, 3D biomedical images consist of subvolumes that represent parts of an \textit{implicit} hierarchical structure. In our case, for any single 3D voxel-grid, we wish to embed and infer the implicit relationships between all of its subvolumes without any supervision. 

% However, many of these data types have an \textit{explicit} tree-like structure encoded in the input data. Complex biomedical images instead consist of subvolumes that capture parts of an \textit{implicit} hierarchical structure from the whole 3D volume. In our case, for any single 3D voxel-grid, we wish to embed and infer the implicit relationships between all of its subvolumes without any supervision. 
% However, in these settings, the objects, and in most cases their relationships, are explicitly encoded in the data.
% In our case, for any single 3D voxel-grid, we wish to embed and infer the relationships between all of its subvolumes without any supervision. 

To embed our 3D images in hyperbolic space, we use a 3D hyperbolic variational autoencoder (VAE). For the decoder, we propose a gyroplane convolutional layer which maps from the latent space back to 3D images while respecting hyperbolic geometry. In addition to the VAE loss, we propose an essential self-supervised loss to capture the hierarchical structure present in the data. More specifically, we consider reconstruction of implicit hierarchies as a pretext task. Concretely, we add a triplet loss which encourages a child subvolume to be mapped close to its parent subvolume in hyperbolic space. To capture hierarchical relationships of varying granularity, we train on subvolumes sampled at multiple scales. Finally, for a specified scale, we cluster the subvolumes in latent space and produce a segmentation map.

%We evaluate our model on three datasets spanning multiple domains: a synthetic dataset, a medical dataset and a biology dataset. 
We evaluate our model on datasets with different domains: synthetic datasets and a medical image dataset. We construct synthetic datasets where we generate structures at various scales and show that our model segments objects at multiple levels of hierarchy better than all prior unsupervised segmentation methods. We demonstrate performance gains ranging from 7\% for the smallest objects to 32\% for the largest objects. On the real-world medical image dataset (BraTS Brain Tumor Segmentation Challenge) \citep{menze2014multimodal, bakas2017advancing,  bakas2018identifying}, we show that our method outperforms prior works by 19\%, and even achieves comparable performance to semi-supervised methods although we do not use any labels. 
% Finally, on the biology dataset (a cryo-ET example), we show the potential for scientific discovery by identifying new sub-cellular objects within 3D images.

\section{Related Work}
\paragraph{Segmentation on 3D voxel data}
Many diverse biomedical images, ranging from MRI and CT scans to fluorescence microscopy, come in the form of 3D voxel-grids. Since 3D voxel-grids are dense, computer vision tasks such as supervised segmentation are commonly performed using deep learning architectures with 3D convolutional layers \citep{chen2016combining, dou20173d, hesamian2019deep, zheng2019new}. However, due to the challenges of obtaining voxel-level annotations in 3D, there has been significant effort in finding semi-supervised approaches, including using labels only from several fully annotated 2D slices of an input volume \citep{cciccek20163d}, using a smaller set of segmentations with joint segmentation and registration \citep{xu2019deepatlas}, and using one segmented input in conjunction with other unlabelled data \citep{zhao2019data}.

Unsupervised approaches for 3D segmentation are useful not only for further reducing the manual annotation effort required, which is especially expensive for segmentation, but also for scientific discovery tasks where we would like to identify previously unknown structures for which annotations are impossible to produce. \cite{moriya2018unsupervised} extends to 3D data an iterative approach of feature learning followed by clustering \citep{yang2016jule}. \cite{nalepa2020hyperspectral} uses a 3D convolutional autoencoder architecture and performs clustering of the latent representations. Another approach, \citep{dalca2018anatomical}, uses a network pre-trained on manual segmentations from a separate dataset to perform unsupervised segmentation of 3D biomedical images. However, this limits applicability to areas where we already have a dataset with manual annotations and makes it unsuitable for unbiased scientific discovery. \cite{gur2019unsupervised} and \cite{kitrungrotsakul2019vesselnet} develop unsupervised methods for 3D segmentation of vessel structures, but these are specialized and do not generalize to the segmentation of other structures, \citep{uzunova2019unsupervised} utilizes knowledge of background patches with no patholoogy, and \cite{baur2018deep} uses deep autoencoding models for unsupervised anomaly detection.

Another line of work performs unsupervised 2D segmentation, such as \cite{ji2019invariant} which proposes a mutual information objective for clustering, and \cite{caron2018deep}, which uses the clustered output of an encoder as pseudo-labels. While these methods can be applied to 2D slices of a 3D volume to perform 3D segmentation, they generally suffer limitations due to insufficient modeling of the 3D spatial information. None of the aforementioned approaches explicitly learn hierarchical structure of the data, which is the main focus of our work.

\paragraph{Hyperbolic representations}
A recent line of work employs hyperbolic space to represent hierarchical structures, with the intuition that tree structures can be naturally embedded into continuous hyperbolic space \citep{nickel2017poincare}. These works utilize hyperbolic representations for a variety of tasks, including MNIST classification \citep{mathieu2019poincare, nagano2019wrapped, ovinnikov2019poincar}, natural language processsing tasks such as embedding word taxonomies and entailment \citep{nickel2017poincare, ganea2018hyperbolicnn}, link prediction and node classification \citep{chami2019hgcn}, and game playing \citep{nagano2019wrapped}. In most of these works, hierarchical structure is \textit{explicitly} encoded in data. In contrast, we seek to capture \textit{implicit} hierarchical structure arising from composition within 3D images.

Several architectures have been proposed in order to learn hyperbolic representations, including hyperbolic VAEs \citep{mathieu2019poincare, nagano2019wrapped, ovinnikov2019poincar}, feed-forward and recurrent hyperbolic neural networks architectures \citep{ganea2018hyperbolicnn}, and hyperbolic graph convolutional networks \citep{chami2019hgcn}. We extend the hyperbolic VAE framework to the task of learning hyperbolic representations from subvolumes of complex 3D images, and use this to perform unsupervised segmentation.
% including a novel hierarchical triplet loss, hyperbolic convolutional layer, and structure-informed sampling scheme to capture relationships among multiple levels of granularity in a given input. 
% We note that our method is general and can also be used for general 3D unsupervised segmentation. \JG{Does this sentence belong here?}

\paragraph{Self-supervision}
Providing self-supervision by solving pretext tasks is one common approach for learning unsupervised visual representations. Pretext tasks leverage properties of the input data or prior knowledge as supervisory signals in order to learn better representations. Examples of pretext tasks include finding the relative position of two patches sampled from an image \citep{doersch2015unsupervised}, solving jigsaw puzzles \citep{noroozi2016unsupervised}, and predicting pixel movements of videos in subsequent frames \citep{pathak2017learning}. 
% In biomedical self-supervision, approaches tend to rely on strong assumption such as longitudinal \citep{jamaludin2017self} or anatomical consistency \citep{bai2019self, chaitanya2020contrastive}. \citet{ouyang2020self} propose superpixel pseudolabels as supervision. 
In contrast, we propose the pretext task of reconstructing implicit hierarchy in 3D voxel-grid images, to learn effective hyperbolic representations for downstream segmentation.

\section{Methods}
In this section, we describe our approach for learning hyperbolic representations of subvolumes (3D patches) from 3D voxel-grid data. We propose a model that comprises a 3D convolutional variational autoencoder (VAE) with hyperbolic representation space and a gyroplane convolutional layer. 
%along with a novel hierarchical triplet loss and a multi-scale sampling scheme as self-supervisory signal that learns implicit hierarchy reconstruction as a pretext task. 
We train our model with self-supervision through a novel hierarchical triplet loss and multi-patch sampling scheme. Then, we cluster the learned representations using hyperbolic $k$-means to produce 3D segmentations. In Section~\ref{prelims}, we provide an overview of hyperbolic space. In Section~\ref{architecture}, we describe our VAE framework with the proposed gyroplane convolutional layer and self-supervised hierarchical triplet loss. Finally, in Section~\ref{segmentation}, we discuss our approach of hyperbolic clustering for segmentation.

\subsection{Hyperbolic formulation}
\label{prelims}
\paragraph{Hyperbolic space}
We embed subvolumes of 3D voxel-grid data in hyperbolic space, a non-Euclidean space with constant negative curvature. In negative curvature spaces, the area of a disc increases exponentially with the radius. We can think of this growth as analogous to the exponential increase of leaves at each level of a tree. Hence hyperbolic space can encode trees with arbitrarily low error \citep{sarkar2011low} and can be considered as the continuous version of hierarchical structures. Unlike trees, hyperbolic space is smooth, permitting our use of deep learning on representations. For additional background on geometry and hyperbolic space, see the Appendix.
%Due to its negative curvature, hyperbolic space can be considered as the continuous version of a tree, making it a natural choice for embedding hierarchical structures. Trees can be embedded in the Poincar\'{e} ball with arbitrarily low error \citep{sarkar2011low}, and like the leaves of a tree, the area of a disc in the Poincar\'{e} ball increases exponentially with the radius. 

\paragraph{Poincar\'{e} ball model of hyperbolic geometry} In this work we use the Poincar\'{e} ball as our model of hyperbolic geometry. The Poincar\'{e} ball (of curvature $c = -1$) is the open ball of radius $1$ centered at the origin equipped with the \textit{metric tensor} $\mathfrak{g}_p = (\lambda_x)^2 \mathfrak{g}_e$, where the conformal factor $\lambda_x = \frac{2}{1 - ||x||^2}$ and $\mathfrak{g}_e$ is the Euclidean metric tensor (i.e., the Euclidean dot product). Formally, this makes the Poincar\'{e} ball a Riemannian manifold. For an introduction to Riemannian manifolds, see the Appendix. The distance $\mathbf{d_p}$ between points on the Poincar\'{e} ball is given by:
\begin{align}\label{eqn:poincaredist}
    \mathbf{d_p}(x, y) = \cosh^{-1}\left(1 + 2 \frac{||x - y||^2}{(1 - ||x||^2)(1 - ||y||^2)} \right)
\end{align}

We use the exponential and logarithm maps to map from Euclidean space to the Poincar\'{e} ball and vice versa. On the Poincar\'{e} ball, we note that the exponential and logarithm maps have the closed form expressions 
\begin{align}\label{eqn:explogmap}
    &\exp_z(v) = z \oplus \left( \tanh \left(\frac{\lambda_z ||v||}{2} \right) \frac{v}{||v||} \right) \\
    &\log_z(y) = \frac{2}{\lambda_z} \tanh^{-1}(||-z \oplus y||)\frac{-z \oplus y}{||-z \oplus y||}
\end{align}
where $\oplus$ denotes Mobius addition, which was introduced by \cite{ungar2001hyperbolic} as a way to define vector operations on hyperbolic space. 

\paragraph{Wrapped Normal Distribution}
The importance of the normal distribution in Euclidean space has led to many generalizations of the normal distribution to Riemannian manifolds. We use the wrapped normal distribution \citep{mathieu2019poincare, nagano2019wrapped}, which can be defined on an arbitrary Riemannian manifold as the push-forward measure obtained by mapping the normal distribution in Euclidean space along the manifold's exponential map. On the Poincar\'{e} ball, the probability density function of the wrapped normal with mean $\mu$ and covariance $\Sigma$ is:
\begin{align}
    \mathcal{N}_{P}(z|\mu, \Sigma) = \mathcal{N}_{E}(\lambda_{\mu}(z)|0, \Sigma) \left( \frac{\mathbf{d_p}(\mu, z)}{\sinh(\mathbf{d_p}(\mu, z))} \right)
\end{align}
where the subscripts $P, E$ indicate distributions over the Poincar\'{e} ball and Euclidean space, respectively. We use the sampling and reparametrization scheme of \cite
{mathieu2019poincare} in order to sample and train our VAE using the wrapped normal distribution. 
% To use the wrapped normal in a VAE, we require both a way to sample from the wrapped normal as well as a way to train its parameters. \citet{mathieu2019poincare} provides a reparametrization and sampling scheme for the wrapped normal on the Poincar\'{e} ball. 

\subsection{Unsupervised hyperbolic representation learning}
\begin{figure*}[ht!]
\vskip 0.2in
\begin{center}
\centerline{\includegraphics[width=1.0\linewidth]{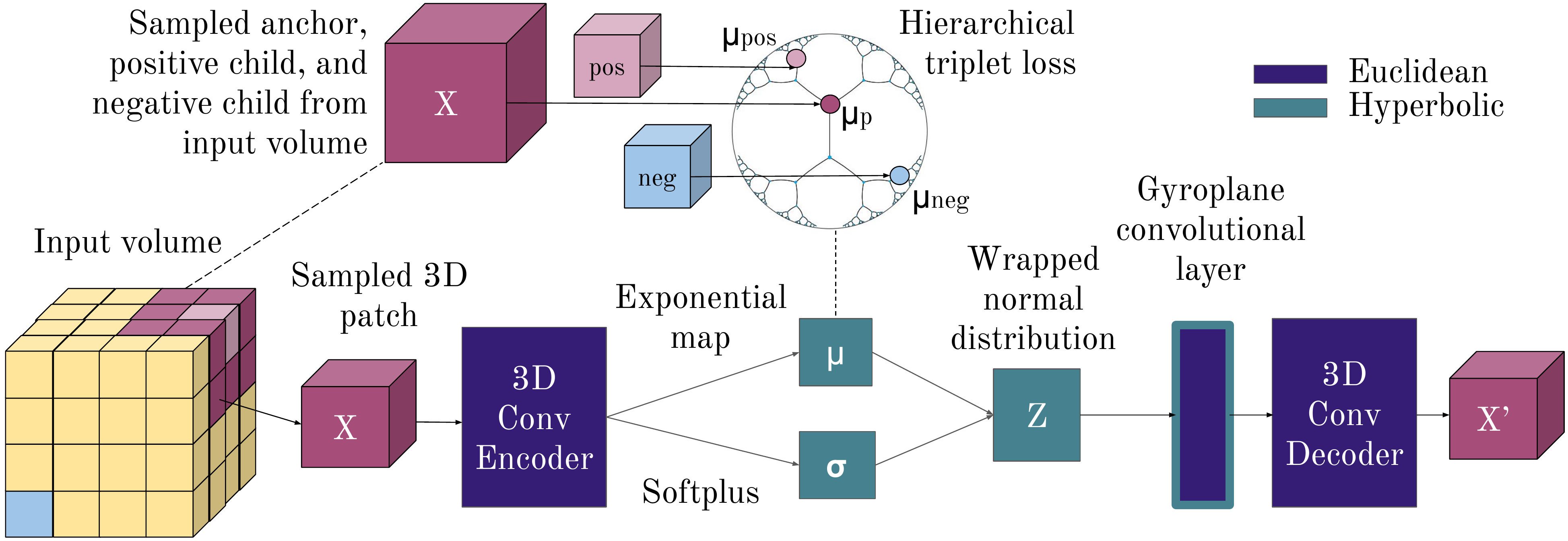}}
\caption{Our method learns hyperbolic representations of subvolumes of 3D voxel-grid data through a 3D hyperbolic VAE with a gyroplane convolutional layer. We enhance the VAE training objective with a self-supervised hierarchical triplet loss that facilitates learning implicit hierarchical structure within the VAE's hyperbolic latent space.}
\label{fig:pull}
\end{center}
\vskip -0.2in
\end{figure*}

\paragraph{3D hyperbolic variational autoencoder}
\label{architecture}
% The VAE framework \citep{kingma2013auto, rezende2014stochastic} is widely used for unsupervised representation learning, but requires new innovations to learn effective hierarchical representations in 3D biomedical images. 
We propose a hyperbolic VAE that consists of a 3D convolutional encoder and decoder to handle 3D input. Our 3D convolutional encoder maps sampled subvolumes of the input into hyperbolic space and produces the parameters of the variational posterior. Our 3D convolutional decoder then reconstructs the 3D subvolumes from sampled latent hyperbolic representations. To ensure that both the encoder and decoder respect the geometry of the latent space, we follow \cite{mathieu2019poincare} and apply an exponential map to the output of the encoder, and use our novel gyroplane convolutional layer as the first layer of the decoder.
%The last layer of the encoder is an exponential map that ensures that the output is in hyperbolic space, and the first layer of the decoder is our proposed gyroplane convolutional layer which maps hyperbolic space to Euclidean space. This ensures that both the encoder and decoder respect the hyperbolic geometry of the latent space. 
We define the prior and variational posterior to be the wrapped normal distribution, which encourages our hierarchical representations to spread out on the Poincar\'{e} ball. Figure \ref{fig:pull} illustrates an overview of our VAE framework. 

Our variational autoencoder takes as input a patch of fixed size $m \times m \times m$. The model learns representations of subvolumes from the input ($X$ in Figure \ref{fig:pull}) that can subsequently be used to perform voxel-level segmentation of the whole volume. To learn hierarchical structure in 3D images, we train the VAE on 3D patches generated using a multi-scale sampling scheme that samples patches of size $r \times r \times r$, where size $r$ is randomly sampled and resized to $m$. Our method learns to embed each patch as part of a hierarchy in hyperbolic space.
%Each sampled patch represents components from the 3D biomedical image that we want to embed into our continuous hierarchy in hyperbolic space.

\paragraph{Gyroplane convolutional layer}
For learning better hyperbolic representations of 3D images, we introduce a gyroplane convolutional layer to effectively map from hyperbolic embedding space to Euclidean space. This allows us to keep the advantages of convolutional layers, such as locality, weight sharing, and translation equivariance. Our model's encoder output ($\mu$ in Figure \ref{fig:pull}) has a product manifold structure, since it is a Cartesian product of vectors in hyperbolic space. To map this to a 3D image in Euclidean space, we generalize the usual Euclidean convolutional layer by replacing the Euclidean affine transformation with an affine transformation on the manifold. 
% Our model's encoder mean output (See $\mu$ in Figure \ref{fig:pull}) can be interpreted as a product of Poincar\'{e} balls, which justifies our definition and use of the gyroplane convolutional layer. 
%Since $\mathbb{R}^n = \mathbb{R} \times \ldots \times \mathbb{R}$, high-dimensional Euclidean spaces can be decomposed into a product of low-dimensional Euclidean spaces. An equivalent decomposition does not hold for arbitrary Riemannian manifolds, making it difficult to generalize the usual (Euclidean) convolutional layer to arbitrary Riemannian manifolds. For manifolds that are products of manifolds, we can generalize the usual convolution by replacing the Euclidean affine transformation with an affine transformation on the manifold. For the Poincar\'{e} ball, one analogue of the Euclidean affine transformation is the gyroplane operator $f_{a ,p}$. 

% For the Poincar\'{e} ball, one analogue of the Euclidean affine transformation is the gyroplane layer $f_{a ,p}$ \citep{ganea2018hyperbolicnn}. 
One way to define an affine transformation on the Poincar\'{e} ball is the gyroplane layer \citep{ganea2018hyperbolicnn}. The derivation of the gyroplane layer is motivated by the fact we can express a Euclidean affine transformation as: $\langle a, z - p \rangle = \sign(\langle a, z - p \rangle) ||a|| \mathbf{d_E}(z, H_{a, p})$ where $\mathbf{d_E}$ is Euclidean distance and $H_{a, p} = \{z \in \mathbb{R}^p | \langle a, z - p \rangle = 0\}$. $H_{a, p}$ is called the decision hyperplane. \cite{ganea2018hyperbolicnn} defines the gyroplane layer $f_{a, p}$ from this formulation by replacing each component with its hyperbolic equivalent: 
\begin{align}\label{eqn:gyroplane}
    f_{a, p}(z) = \sign \left( \langle a, \log_p(z) \rangle_p \right) |a|_p \mathbf{d_p}(z, H_{a, p})
\end{align}
where $H_{a, p}$ is the hyperbolic decision boundary $H_{a, p} = \{ z \in \mathcal{B} | \langle a, \log_p(z) \rangle = 0 \}$, and the distance to the hyperbolic decision boundary $\mathbf{d_p}(z, H_{a, p})$ is
\begin{align}
    \mathbf{d_p}(z, H_{a, p}) = \sinh^{-1} \left( \frac{2|\langle -p \oplus z, a \rangle|}{(1 - ||-p \oplus z||^2)||a|} \right)
\end{align}

We can now define our gyroplane convolutional layer by generalizing the Euclidean affine transformation using the gyroplane layer. For simplicity, suppose $x$ is a 4D tensor containing elements of the Poincar\'{e} ball and our kernel size is $k \times k \times k$, with an odd $k$ value. Our gyroplane convolutional layer is defined as:
\begin{align}\label{eqn:gyroconv}
    y_{r, s, t} = \sum_{\alpha = r - \lfloor k/2 \rfloor}^{r + \lfloor k/2 \rfloor} \sum_{\beta = s - \lfloor k/2 \rfloor}^{s + \lfloor k/2 \rfloor} \sum_{\gamma = t - \lfloor k/2 \rfloor}^{t + \lfloor k/2 \rfloor} f_{a, p}(x_{\alpha, \beta, \gamma})
\end{align}
The gyroplane convolutional layer can be extended in the same way as Euclidean convolutional layers to incorporate even kernel size $k$, input and output channels, padding, stride, and dilation. 

\paragraph{Self-supervised hierarchical triplet loss}
As our model is trained on subvolumes of the 3D input, we cannot easily obtain the \textit{implicit} hierarchical structure of the whole volume.
%[TODO: this is too strong, any given forward pass, no hierarchical signal here] from any single model input. 
% As our model is trained on patches of the whole 3D volume, the hierarchical structure of the volume is not readily apparent from the individual inputs.
% We introduce self-supervision in the form of a hierarchical triplet loss for the model to learn to infer \textit{implicit} hierarchical structure as a pretext task, with a new multi-patch sampling scheme that encourages parent-child relationships to form on the Poincar\'{e} ball.
To encode this structure in our model, we introduce self-supervision through the reconstruction of inferred hierarchy as a pretext task. This task encourages our learned representations on the Poincar\'{e} ball to reflect parent-child relationships of the input's implicit hierarchy.
%This task encourages capturing parent-child relations of the implicit hierarchy on the Poincar\'{e} ball. \JH{think about later} %To capture the diversity of the 3D image, we employ a multi-patch sampling scheme.

% We propose a triplet loss that serves as a hierarchy reconstruction pretext task. 
Our self-supervision takes the form of a triplet loss that encourages hierarchically-related patches to have more similar representations. Since any two arbitrary patches may have some hierarchical relationship, we sample patches for our triplet loss to capture hierarchy in a tractable way. To sample 3D patches for our triplet loss, we first sample an anchor patch that acts as our parent patch (red volume $X$ in Figure \ref{fig:pull}). We then sample the positive patch as a smaller subpatch that resides within the anchor patch (pink volume $pos$ in Figure \ref{fig:pull}). The anchor and positive patches form a parent-child relationship that we encourage to have closer representations in hyperbolic space, which has the interpretation as belonging to the same branch of the hierarchy ($\mu_{p}$ and $\mu_{\mathrm{pos}}$ in Figure \ref{fig:pull}). The exponentially growing surface area near the edge of hyperbolic space allow this natural parent-child structure to form. We then sample a negative patch as a spatially distant patch (blue volume $neg$) that does not overlap with the anchor patch. The triplet loss encourages the the negative patch's representation to belong to a different branch of the hierarchy ($\mu_{p}$ and $\mu_{\mathrm{neg}}$).

% To sample 3D patches for the triplet loss, we first generate an anchor patch, our parent, centered at voxel $v$ with size $r \times r \times r$ according to one of the above sampling schemes. A positive child patch is generated as a smaller sub-patch of the anchor patch as follows: the positive child patch is centered at $v$ with size $r_\mathrm{child} \times r_\mathrm{child} \times r_\mathrm{child}$, where $r_\mathrm{child} \sim \mathcal{U}(\ell_\mathrm{min}, r - r_\mathrm{gap})$, and $r_\mathrm{gap}$ is a hyperparameter representing the gap in size between the anchor size and the child size. The anchor patch and positive child represents a parent-child relationship that we implicitly encourage to lie closer in representational space; in hyperbolic space, the exponentially growing surface area near the edges allow for this natural parent-child structure to form. 

% In addition, a negative child is a patch of size $r_\mathrm{child} \times r_\mathrm{child} \times r_\mathrm{child}$ centered at $v_\mathrm{neg}$, where $v_\mathrm{neg}$ is sampled uniformly from the set of voxels $w$ such that a patch of size $r_\mathrm{child} \times r_\mathrm{child} \times r_\mathrm{child}$ centered at $w$ does not overlap with the anchor patch. The negative child serves as a semantically more dissimilar representation than the positive child in our contrastive loss formulation. 

% [sampling move to experiment, key is parent child here, in VAE section key is subvolumes, add more intuition as self-supervised, pretext task is key idea!!]

Our choice of positive and negative patches is motivated by the compositional hierarchy of 3D volumes. The hierarchical triplet loss encourages the anchor patch and a sub-patch (parent and positive child) to have similar representations, while encouraging the anchor patch and a distant patch (parent and negative child) to have dissimilar representations. %In hyperbolic space, this has the interpretation of belonging to one branch of the hierarchy on the Poincar\'{e} ball and belonging to a different branch, respectively.
We use this implicit hierarchy reconstruction as a pretext task to encourage learning relationships between nested objects in 3D biomedical images. Our multi-patch sampling scheme and triplet loss formulation allows representations to encode \textit{implicit} structure in hyperbolic space.

Our hierarchical triplet loss can be formulated with any dissimilarity measure $\mathbf{d}$ between the encoder outputs $\mu$ (see Figure~\ref{fig:pull}) of the anchor $\mu_\mathrm{p}$, positive child $\mu_\mathrm{pos}$, and negative child $\mu_\mathrm{neg}$. For our model, we take $\mathbf{d}$ to be the Poincar\'{e} ball distance $\mathbf{d_p}$ and define our triplet loss with margin $\alpha$ as:
\begin{align}\label{eqn:tripletloss}
    L_\mathrm{triplet}(&\mu_\mathrm{p}, \mu_\mathrm{pos}, \mu_\mathrm{neg}) := \max(0, \mathbf{d_p}(\mu_\mathrm{p}, \mu_\mathrm{pos}) - \mathbf{d_p}(\mu_\mathrm{p}, \mu_\mathrm{neg}) + \alpha)
\end{align}
This formulation can be extended to any metric space by taking the dissimilarity measure $\mathbf{d}$ to be the space's metric. In particular, for our ablations using an Euclidean latent space, we take the dissimilarity measure $\mathbf{d}$ to be the Euclidean distance. 

\paragraph{Optimization} We optimize a loss function that can be decomposed as the standard evidence lower bound (ELBO) loss for variational autoencoders and our hierarchical triplet loss that encourages the learning of structure in hyperbolic space. \cite{mathieu2019poincare} generalized the ELBO loss to Riemannian manifold latent spaces as
\begin{align}\label{eqn:elbo}
    L_{ELBO} &:= \int_{\mathcal{M}} \log \left(\frac{p_{\theta}(x|z)p(z)}{q_{\phi}(z|x)} \right)q_\phi(z|x) d\mathcal{M}(z) \le \log p(x)
\end{align}
where $d\mathcal{M}(z) = \sqrt{|G(z)|}dz$ is the measure induced on the manifold by the Riemannian metric $G(z)$ (see Appendix). Our total loss is then formulated as
\begin{align}
L_\mathrm{total} = L_\mathrm{ELBO} + \beta L_\mathrm{triplet}
\end{align}
where $\beta$ is a hyperparameter that controls the strength of the hierarchical triplet loss.

\subsection{Segmentation by clustering representations} 
\label{segmentation}
\paragraph{Hyperbolic clustering} In 3D segmentation, we seek to assign each voxel $v$ a segmentation label $s_\mathrm{v} \in \{1, \ldots, n\}$, where $n$ is the number of segmentation classes. We perform segmentation by clustering the representations of patches centered at each voxel. We first use our trained VAE encoder to generate latent representations $\mu_\mathrm{v}$ for each voxel $v$. We do this by taking a patch of fixed size $p \times p \times p$ centered at $v$, upsampling or downsampling it to the encoder input size $m \times m \times m$, and then encoding the patch to retrieve $\mu_\mathrm{v}$. We then cluster the $\mu_v$ into $n$ clusters, and produce a segmentation by assigning each $v$ the cluster label of $\mu_v$. We perform clustering through a $k$-means algorithm that respects hyperbolic geometry, which we derive by replacing the Euclidean centroid and distance computations of classical $k$-means with their counterparts in Riemannian geometry, the Fr\'{e}chet mean and manifold distance. We calculate the Fr\'{e}chet mean using the algorithm of \cite{lou2020differentiating}.

\section{Experiments}
\label{experiments}
% Though our method could be applied to any 3D voxelized grid data, we evaluate on several biomedical datasets due to the availability of annotated 3D voxel data in the field. 
We evaluate our method quantitatively on both synthetic 3D datasets simulating biological image data as well as the real-world Brain Tumor Segmentation (BraTS) tumor segmentation dataset. Our biologically-inspired synthetic datasets allows quantitative evaluation of segmentation at multiple levels of hierarchy, while the BraTS dataset is a well-known benchmark for 3D MRI segmentation. 
% We also demonstrate the use of unsupervised segmentation for discovering new biological structures on an example of real-world cryogenic electron tomography (cryo-ET) data. 
%However, our method is general and could be applied to any 3D voxelized grid data, not just biomedical images. 

\paragraph{Implementation details}
For all models, the encoder of our variational autoencoder is comprised of four 3D convolutional layers with kernel size $5$ of increasing filter depth $\{16, 32, 64, 128\}$. The decoder has the same structure, except with decreasing filter depth and a gyroplane convolutional layer as the initial layer. We use $\beta = 1e3$ as the weighting factor between $L_\mathrm{ELBO}$ and $L_\mathrm{triplet}$ and $\alpha = 0.2$ as the triplet margin. In all experiments, we fix the representation dimension to be $d = 2$, and show latent dimension ablations in the Appendix. We train our model using the Adam optimizer \citep{kingma2014adam}. For inference, we obtain the latent representations of $5 \times 5 \times 5$ patches densely across the full volume, and then perform hyperbolic $k$-means clustering, where the number of clusters $k$ is a hyperparameter that controls the granularity of the segmentation. For quantitative evaluation, we then use the Hungarian algorithm \citep{kuhn1955hungarian} to match each predicted segmentation class with a corresponding ground truth label.

We utilize two anchor patch sampling schemes, one for input of smaller sizes and one for larger sizes. In both schemes, for a given 3D volume, we sample $i$ patch centers $v_i$ uniformly with patch size $r$, and upsampling or downsampling to size $m \times m \times m$. In the sampling scheme for smaller inputs, the patch size $r$ is sampled uniformly, whereas in the sampling scheme for larger inputs, $r$ is sampled log-uniformly. This scheme is motivated by the following observations: for larger patches, a small change in $r$ is less likely to correspond to significant semantic difference, and inherent structure causes the different levels of hierarchy to naturally follow a log scale. For training on the synthetic dataset, we sample 3D volume sizes uniformly, and for BraTS we sample using the log scale.
%For training on the synthetic dataset, we sample 3D volume sizes uniformly, and for BraTS and the cryo-ET dataset we sample using the log scale.

For every sample of an anchor patch of size $r \times r \times r$, we generate a positive child patch as a smaller patch of the anchor patch as follows: the positive child patch is a subvolume within the anchor patch with size $r_\mathrm{child} \times r_\mathrm{child} \times r_\mathrm{child}$, where $r_\mathrm{child} \sim \mathcal{U}(\ell_\mathrm{min}, r - r_\mathrm{gap})$, and $r_\mathrm{gap}$ is a hyperparameter representing the gap in size between the anchor size and the child size. The negative child is a patch of size $r_\mathrm{child} \times r_\mathrm{child} \times r_\mathrm{child}$ that does not overlap with the anchor patch.

\subsection{Biologically-inspired synthetic dataset}
\label{toy}

\begin{figure}[htb!]
\vskip 0.2in
\begin{center}
\centerline{\includegraphics[width=0.6\columnwidth]{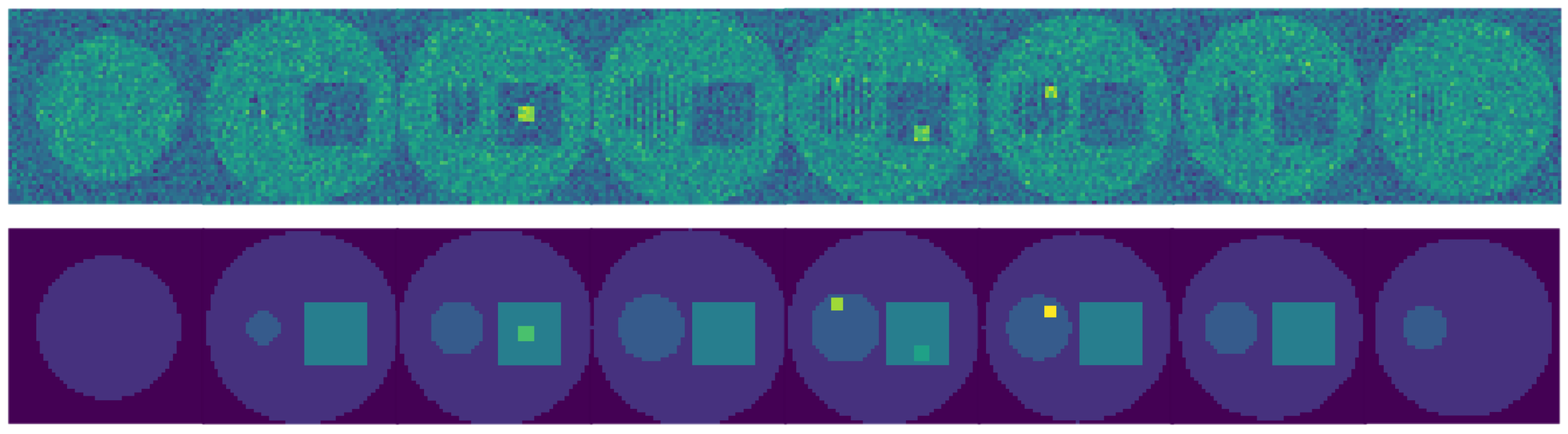}}
\caption{Sampled 2D slices from a 3D volume in our biologically-inspired synthetic dataset. The top row showcases the raw input data, and the bottom row showcases the ground truth segmentation.}
\label{fig:toy-dset-examples}
\end{center}
\vskip -0.2in
\end{figure}

Since we want to evaluate segmentation performance at multiple levels of hierarchy and most 3D datasets do not have the necessary annotation, we first generate a synthetic dataset. This dataset enables a more thorough evaluation of the effectiveness of our model for unsupervised 3D segmentation. Our synthetic dataset is inspired by cryo-ET images of cells. Each volume in our synthetic dataset contains multiple levels of hierarchy with the objects at each level differentiated by texture, size, and shape. Figure~\ref{fig:toy-dset-examples} shows an example input volume with sampled slices shown. Our dataset consists of $120$ total volumes, which we split into $80$ training, $20$ validation, and $20$ test examples. Each synthetic volume has size $50 \times 50 \times 50$. Additional information on the synthetic dataset generation process as well as a more difficult version of the dataset, where the boundaries of each shape are perturbed, is described and benchmarked in Appendix A.5.

\vspace{-0.4cm}

To demonstrate segmentation performance on objects at different scales, we evaluate on the three levels of hierarchy defined above and use the average class DICE score to compare segmentation performance. Since our model is unsupervised, we assign segmentation classes to ground truth labels using the Hungarian algorithm. See results in Table~\ref{table:toy-comparison} and Table~\ref{table:ablation-table}. We also show results on a more challenging irregular synthetic dataset in Appendix A.5.

\begin{table*}[htb!]
\caption{Comparison with prior approaches on synthetic dataset}
\vskip 0.15in
\begin{center}
\begin{small}
\begin{sc}
\begin{tabular}{llllll}
    \toprule
    & Dice \textit{Level 1} & Dice \textit{Level 2} & Dice \textit{Level 3} & Supervision type \\
    \midrule \citet{cciccek20163d} &
 0.968 & 0.829 & 0.668 & 3D Semi-supervised \\ \citet{zhao2019data} & 0.989 & 0.655 & 0.357 & 3D Semi-supervised \\
    \midrule \citet{nalepa2020hyperspectral} & 0.530 & 0.276 & 0.112 & 3D Unsupervised \\ \citet{ji2019invariant} & 0.589 & 0.291 & 0.150 & 2D to 3D Unsupervised \\ \citet{moriya2018unsupervised} & 0.628 & 0.311 & 0.141 & 3D Unsupervised  \\
    % \citet{ouyang2020self} & 0.744 & 0.528 & \textbf{0.274} & 3D Unsupervised \\
    \textbf{Ours} & \textbf{0.952} & \textbf{0.541} & \textbf{0.216} & 3D Unsupervised \\
    \bottomrule
  \end{tabular}
\end{sc}
\end{small}
\end{center}
\vskip -0.1in
\label{table:toy-comparison}
%\vspace{-0.4cm}
\end{table*}

\begin{table*}[htb!]
\caption{Ablation studies on synthetic dataset}
\vskip 0.15in
\begin{center}
\begin{small}
\begin{sc}
  \begin{tabular}{llllll}
    \toprule
    Latent Space & Configuration & Dice \textit{Level 1} & Dice \textit{Level 2} & Dice \textit{Level 3}  \\
    \midrule
    Euclidean & Base & 0.784 & 0.322 & 0.109 \\
     & Triplet & 0.761 & 0.342 & 0.153 \\
    \midrule
    Hyperbolic & Base & 0.832 & 0.352 & 0.135 \\
    & GyroConv & 0.905 & 0.473 & 0.204\\
     & Triplet & 0.945 & 0.534 & \textbf{0.222} \\
     & GyroConv \& Triplet & \textbf{0.952} & \textbf{0.540} & 0.216 \\
    \bottomrule 
  \end{tabular}
\end{sc}
\end{small}
\end{center}
\vskip -0.1in
\label{table:ablation-table}
% \vspace{-0.4cm}
\end{table*}

\paragraph{Comparison with prior approaches}
Table~\ref{table:toy-comparison} shows quantitative comparison of our method with prior state-of-the-art 3D unsupervised and 2D unsupervised (which we extend to 3D) models. In addition, we also compare our method to prior semi-supervised work, as unsupervised 3D segmentation is a relatively unexplored field, and we provide baselines with different levels of supervision for additional reference. \citet{cciccek20163d} was trained with $2\%$ of the ground truth slices in each of the $xy$, $yz$, and $xz$ planes, and \citet{zhao2019data} was trained with one fully annotated atlas, which can both still be expensive given the large size of many 3D biomedical images. \citet{ji2019invariant} was implemented using the authors' original code and extrapolated to 3D by applying the method to each slice. For \citet{nalepa2020hyperspectral} and \citet{moriya2018unsupervised}, we re-implemented their methods as the original code was unavailable. Our model performs significantly better than all unsupervised prior work at all levels of hierarchy. We also perform comparably to the semi-supervised approach of \citet{zhao2019data}, despite not using any labels.

\paragraph{Ablation}
Table~\ref{table:ablation-table} presents ablation studies on the hierarchical synthetic dataset comparing our contributions: Euclidean vs. hyperbolic representations, the addition of our gyroplane convolutional layer, and the addition of our hierarchical triplet loss. The base Euclidean configuration is the 3D convolutional VAE with Euclidean latent space, no gyroplane convolutional layer, and trained with just the ELBO loss. The triplet Euclidean configuration adds the hierarchical triplet loss to the base Euclidean configuration. The base hyperbolic configuration is the same as the base Euclidean configuration except with hyperbolic latent space. The triplet configuration is the hyperbolic analogue of the Euclidean triplet configuration, and gyroconv configurations have the addition of the gyroplane convolutional layer. 

Hyperbolic representations outperform their Euclidean counterparts in all experiments. We attribute this to the more efficient and better organization of hyperbolic representations. When we introduce our self-supervised triplet loss, performance improves significantly for our hyperbolic models. Performance for our Euclidean model does not improve as much, likely due to information loss in representing hierarchical input. Introducing the gyroplane convolutional layer shows clear improvement over the base hyperbolic model, which demonstrates the benefit of having a convolutional layer that respects the geometry of the latent space. The combination of the triplet loss and gyroplane convolutional layer exhibits the most gain over the base hyperbolic model, with smaller gains over the model with just the added triplet loss. We show the importance of our hierarchical self-supervision for learning effective representations that capture implicit hierarchical structure.

\begin{table*}[t]
\caption{Table shows comparison on BraTS 2019 dataset. Figure shows a qualitative example where top left image is a slice from a 3D test volume, and the three other images show segmentations with $2, 3, 4$ numbers of clustering centroids respectively, illustrating multiple levels of hierarchy learned.}
\vskip 0.2in
\begin{center}
\begin{small}
\begin{sc}
\begin{minipage}[b]{0.6\linewidth}
\centering
  \begin{tabular}{lllll}
    \toprule
    BraTS dataset & Dice WT & Algorithm type \\
    \midrule
    SOTA \citep{jiang2019two} & 0.888 & 3D Fully-sup. \\
    \midrule \citet{zhao2019data} & 0.648 & 3D Semi-sup. \\ \citet{cciccek20163d} & 0.760 & 3D Semi-sup. \\
    \midrule \citet{ji2019invariant} & 0.211 & 2D-to-3D Unsup. \\ 
    % \citet{ouyang2020self} & 0.231 & 3D Unsupervised \\
    \citet{moriya2018unsupervised} & 0.425 & 3D Unsup.  \\ \citet{nalepa2020hyperspectral} & 0.495 & 3D Unsup. \\
    \textbf{Ours} & \textbf{0.684} & 3D Unsup. \\
    \bottomrule
  \end{tabular}
\end{minipage}\hfill
\begin{minipage}{0.4\linewidth}
\centering
\includegraphics[width=4.0cm]{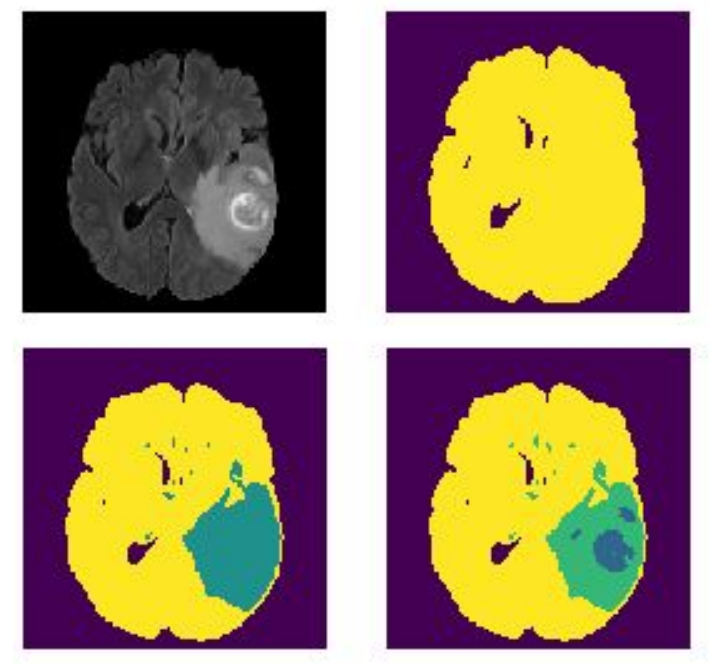}
\end{minipage}
\end{sc}
\end{small}
\end{center}
\vskip -0.1in
\label{table:brats-table}
%\vspace{-0.4cm}
\end{table*}

%\begin{figure*}[ht]
%\vskip 0.2in
%\begin{center}
%\centerline{\includegraphics[width=14cm]{figures/cryo-paper-trip.pdf}}
%\caption{Leftmost image is a partial slice from a 3D cryo-ET image. The middle box shows segmentation from our best hyperbolic model, the rightmost box shows segmentation from our best Euclidean model. The segmentations in each box correspond to clustering using 2 vs. 4 classes.}
%\label{fig:cryoem-figure}
%\end{center}
%\vskip -0.2in
%\end{figure*}

\subsection{Brain Tumor Segmentation challenge dataset}
\label{brats_section}
The BraTS 2019 dataset is a public, well-established benchmark dataset containing 3D MRI scans of brain tumors and voxel-level ground truth annotations of tumor segmentation masks \citep{menze2014multimodal, bakas2017advancing,  bakas2018identifying}. The scans are of dimension $200 \times 200 \times 155$ and have four modalities; we use the FLAIR modality, which is the most commonly used one-modality input. We use the same evaluation metric as in the BraTS challenge, and compare DICE score on whole tumor (WT) segmentation, which is detectable solely from FLAIR, as well as average and 95 percentile Hausdorff distance for the competing unsupervised methods (see Table~\ref{table:hausdorff}). There are $259$ high grade glioma (HGG) labelled training examples, which we split into $180$ train, $39$ validation, and $40$ test examples. We do not use the official validation or test sets because the ground truth annotations for these sets are not publicly available.

\begin{wraptable}{r}{7.0cm}
\caption{Comparison of our method against prior unsupervised work in Hausdorff distance. (Lower is better.)}
\vskip 0.15in
\begin{center}
\begin{small}
\begin{sc}
\begin{tabular}{lll}
    \toprule
    & Average & 95\% \\
    \midrule
        \citet{moriya2018unsupervised} & 118.144 & 170.434 \\
        \citet{ji2019invariant} & 96.865 & 114.400  \\
        \cite{nalepa2020hyperspectral} & 87.704 & 110.803 \\
    \textbf{Ours} & \textbf{77.940} & \textbf{97.641} \\
    \bottomrule
  \end{tabular}
\end{sc}
\end{small}
\end{center}
\vskip -0.1in
\label{table:hausdorff}
\end{wraptable}

Table~\ref{table:brats-table} shows the DICE score comparison of our results against prior work. For fair comparison, all baselines are trained with only the FLAIR modality. The only exception is the current state-of-the-art fully-supervised result \citep{jiang2019two} in Table~\ref{table:brats-table}, which uses all 4 modalities. We show this for reference as an upper bound; the reported number is trained on the full train set and evaluated on the BraTS test set. 

Our best model performs significantly better than all prior unsupervised methods, and in addition outperforms one 3D semi-supervised model. This demonstrates the ability of our self-supervised hyperbolic representations to effectively capture the hierarchical structure in individual brain scans. We use a granular segmentation with three clusters for quantitative evaluation in order to capture the tumor, brain, and background, then use the Hungarian algorithm for assignment. In Table \ref{table:hausdorff}, we demonstrate that our model also outperforms all prior methods on average and 95 percentile Hausdorff metrics. In addition, we also show qualitative results for our model (see Figure~\ref{table:brats-table}), which include byproduct segmentations from the same model with different numbers of clusters specified, showcasing additionally discovered features in the scan that could be clinically useful.

%\subsection{Cryogenic electron tomography data}
%\label{cryoem_section}

%Finally, we show a real-world example where unsupervised discovery of new biological organelles is important. Cryogenic electron tomography (cryo-ET) is a technique that images cells at cryogenic temperatures with a beam of electrons. The value of each voxel is the electron density at that location.
%, and is created through reconstruction from tilt slices of $\pm 70$ degrees from electron tomography. 
%Cryo-ET images are a rich source of biological data, capturing many unknown subcellular objects that we would like to identify and understand.

%We train our model on three $512 \times 512 \times 250$ cryo-ET tomograms of cells collected from a research laboratory, and run inference on a fourth tomogram. Figure~\ref{fig:cryoem-figure} shows segmentations produced by our model on a mitochondria from the evaluation tomogram, using our proposed hyperbolic model vs. Euclidean model, and at a coarse and finer level of granularity. Unlike the Euclidean approach, our hyperbolic approach discovers a fine-grained class corresponding to small features on the mitochondria, which may be macromolecular aggregates. We can now investigate the discovered features for their biological identities and functions, leading to greater scientific understanding. 

\section{Conclusion}
We propose a method for learning hyperbolic representations of 3D voxel-grid images that captures the \textit{implicit} hierarchical structure in biomedical data, and show that these representations are well suited for the task of unsupervised 3D segmentation. We conduct our representation learning through a hyperbolic 3D convolutional VAE with a novel gyroplane convolutional layer that respects hyperbolic geometry. We then enhance the VAE training objective with a self-supervised hierarchical triplet loss that learns \textit{implicit} hierarchical structure within the VAE’s hyperbolic latent space as a pretext task. We demonstrate that hyperbolic clustering of learned voxel-level representations can be used to achieve state-of-the-art unsupervised 3D segmentation on synthetic hierarchical datasets and the real-world BraTS dataset. 
% We also illustrate the promise of our approach for unsupervised scientific discovery on cryogenic electron tomography data. 

\textbf{Acknowledgments.} We thank support from the Chan-Zuckerberg Initiative (Neurodegeneration Challenge Network Collaborative Pairs grant DAF2021-221728 to J.H., J.G., G.W., W.C., and S.Y.) and National Institutes of Health (grant number P01NS092525 to W.C.).

\bibliographystyle{abbrvnat}
\bibliography{neurips_2021}

\newpage

%\appendix

%\section{Appendix}

%Optionally include extra information (complete proofs, additional experiments and plots) in the appendix.
%This section will often be part of the supplemental material.

\end{document}

% \maketitle
\appendix

\section{Appendix}

\subsection{Broader impact}
Our work introduces a general method for unsupervised 3D segmentation that can be used for any 3D voxel-grid data. This line of work is especially useful for analyzing biomedical data, as many different types of biomedical data are in volumetric form and lack the ground truth annotations required for fully- or semi-supervised segmentation. For example, we may wish to study diseased tissue but do not have sufficient understanding to ensure that unexplored features of interests are labelled in training data. We illustrate the potential of our proposed approach for scientific discovery applications using our example of cryo-ET data in the Appendix. The discovered features can now be analyzed for their chemical identities and functions, in diseased vs. healthy cells. Similarly, unsupervised discovery of substructures can also enable richer analysis of other types of biomedical data such as CT and MRI scans. Potential negative societal impact of our work could arise from malicious intent in extracting information from certain types of 3D voxel-grid data for ill use, such as data mining from 3D scenes of sensitive domains without consent, which our method facilitates easily without labels. However, we hope our work is utilized to enable new downstream applications primarily from real-world 3D biomedical images, which are one of the most common types of 3D voxel-grid data. 

\subsection{Riemannian manifolds}
\label{appendix:riemmanian}
In this section, we give a more complete introduction to Riemannian manifolds, of which hyperbolic space is an example. Riemannian manifolds are spaces that locally resemble Euclidean space. To define this mathematically, we first introduce a \textit{manifold} as a set of points $\mathcal{M}$ that locally resembles the Euclidean space $\mathbb{R}^n$. Associated with each point $\mathbf{x} \in \mathcal{M}$ is a vector space called the \textit{tangent space} at $\mathbf{x}$, denoted $\mathcal{T}_\mathbf{x} \mathcal{M}$, which is the space of all directions a curve on the manifold $\mathcal{M}$ can tangentially pass through point $\mathbf{x}$. A metric tensor $\mathfrak{g}$ defines an inner product $\mathfrak{g}_\mathbf{x}$ on every tangent space, and a \textit{Riemannian manifold} is a manifold $\mathcal{M}$ together with a metric tensor $\mathfrak{g}$. For each tangent sapce $\mathcal{T}_x \mathcal{M}$, the metric tensor has \textit{matrix representation} $G$ defined as $\mathfrak{g}_{\mathbf{x}}(u, v) = u^TG(\mathbf{x})v$.

Distance on a Riemannian manifold as can defined as the following. Let $\gamma: [a, b] \to \mathcal{M}$ be a curve on the manifold $\mathcal{M}$. The \textit{length} of $\gamma$ is defined to be $\int_a^b |\gamma'(t)|_{\gamma(t)} dt$ and denoted $L(\gamma)$. The \textit{distance} between any two points $\mathbf{x}, \mathbf{y}$ on the manifold is defined as $d_{\mathcal{M}}(\mathbf{x}, \mathbf{y}) = \inf L(\gamma)$, where the $\inf$ is taken over all curves $\gamma$ that begin at $\mathbf{x}$ and end at $\mathbf{y}$. This distance makes $\mathcal{M}$ a metric space. % A \textit{geodesic} is defined as the shortest possible curve between two points. We can also define a measure $d\mathcal{M}(x) = \sqrt{|G(x)|}dx$.
% $\mathfrak{g}_\mathbf{x}$ also defines a norm $|v|_\mathbf{x} = \sqrt{\mathfrak{g}_\mathbf{x}(v, v)}$ on $\mathcal{T}_\mathbf{x} \mathcal{M}$. $\mathfrak{g}_\mathbf{x}$ has a \textit{matrix representation} $G(x)$, which is defined so that $\mathfrak{g}_\mathbf{x}(u, v) = u^TG(x)v$.
%  where $n$ is the \textit{dimension} of manifold $\mathcal{M}$

%We can use exponential and logarithmic maps to traverse between Euclidean space and the aforementioned Riemannian manifold. 
The \textit{exponential map} $\exp_{\mathbf{x}}(v): \mathcal{T}_\mathbf{x} \mathcal{M} \to \mathcal{M}$ is a useful way to map vectors from the (Euclidean) tangent space to the manifold. The exponential map is defined as $\exp_{\mathbf{x}}(v) = \gamma(1)$, where $\gamma$ is the unique geodesic, the shortest possible curve between two points, starting at $\mathbf{x}$ with starting direction $v$. Intuitively, one can think of the exponential map as telling us how to travel one step starting from a point $\mathbf{x}$ on the manifold in the $v$ direction. The logarithmic map $\log_v(x): \mathcal{M} \to \mathcal{T}_\mathbf{x} \mathcal{M}$ is the inverse of the exponential map, and maps vectors back to Euclidean space.

%Additionally, we define mobius addition on the Riemannian manifold as

%$$
%    z \oplus_c y = \frac{(1 + 2c \langle z, y\rangle + c \norm{y}^2)z + (1 - c \norm{z}^2)y}{1 + %2c \langle z, y\rangle + c^2 \norm{z}^2 \norm{y}^2}
%$$

\subsection{Gyrovector operations in the Poincar\'{e} Ball}

Gyrovector operations were first introduced by \cite{ungar2008gyrovector} to generalize the Euclidean theory of vector spaces to hyperbolic space. Mobius addition is the Poincar\'{e} ball analogue of vector addition in Euclidean spaces. The closed-form expression for Mobius addition on the Poincar\'{e} ball with negative curvature $c$ is \cite{mathieu2019poincare}:
\begin{align}\label{mobiusadd}
    z \oplus_c y = \frac{(1 + 2c \langle z, y \rangle + c||y||^2)z + (1 - c||z||^2)y}{1 + 2c\langle z, y \rangle + c^2||z||^2||y||^2}
\end{align}
As one might anticipate, when $c = 0$ we recover Euclidean vector addition. Additionally, the analogue of Euclidean vector subtraction is Mobius subtraction, which is defined as $x \ominus_c y = x \oplus_c (-y)$, and the analogue of  Euclidean scalar multiplication is Mobius scalar multiplication, which can be defined for a scalar $r$ as \citep{ganea2018hyperbolicnn}:
\begin{align}
    r \otimes_c x = \frac{1}{\sqrt{c}}\tanh(r \tanh^{-1}(\sqrt{c}||x||))\frac{x}{||x||}
\end{align}
where we also recover Euclidean scalar multiplication when $c = 0$. In this paper, we only consider the Poincar\'{e} ball with fixed constant negative curvature $c = 1$, which allows us to drop the dependence on $c$.

\begin{table*}[htb!]
\caption{Ablation study of latent space dimension for Euclidean and Hyperbolic models on the synthetic dataset. Dice scores for all three levels are reported.}
\vskip 0.15in
\begin{center}
\begin{small}
\begin{sc}
  \begin{tabular}{lllllll}
    \toprule
    Latent Space & Dice \textit{Level} & d=2 & d=3 & d=5 & d=8 & d=16  \\
    \midrule
    Hyperbolic & \textit{Level 1} & 0.952 & 0.959 & 0.956 & 0.942 & 0.954\\
     & \textit{Level 2} & 0.541 & 0.538 & 0.550 & 0.529 & 0.541 \\
     & \textit{Level 3} & 0.216 & 0.213 & 0.219 & 0.226 & 0.228 \\
    \midrule
    Euclidean & \textit{Level 1} & 0.761 & 0.838 & 0.847 & 0.871 & 0.872 \\
    & \textit{Level 2} & 0.342 & 0.362 & 0.378 & 0.481 & 0.495 \\
    & \textit{Level 3} & 0.153 & 0.176 & 0.165 & 0.225 & 0.228 \\
    \bottomrule 
  \end{tabular}
\end{sc}
\end{small}
\end{center}
\vskip -0.1in
\label{table:ldim-ablation-table}
\end{table*}

\subsection{Latent dimension ablation}
For all sections in our paper, our experiments were all run with latent dimension of $2$. To show the effect of higher latent space dimensions, we report an ablation study for both hyperbolic and Euclidean representations (See Table~\ref{table:ldim-ablation-table}). As expected, for our Euclidean latent space model, performance increases with dimension. However, our hyperbolic model still outperforms the Euclidean model at all tested dimensions, and shows that we can embed representations efficiently at lower dimensions.  

% \begin{table}[htb]
%   \caption{Ablation study of latent space dimension for Euclidean and Hyperbolic models on the synthetic dataset. Dice scores for all three levels are reported.}
%   \centering
%   \begin{tabular}{lllllll}
%     \toprule
%     Latent Space & Dice \textit{Level} & d=2 & d=3 & d=5 & d=8 & d=16  \\
%     \midrule
%     Hyperbolic & \textit{Level 1} & 0.95211 & 0.95943 & 0.95574 & 0.94159 & 0.95350\\
%      & \textit{Level 2} & 0.54065 & 0.53827 & 0.54959 & 0.52889 & 0.54097 \\
%      & \textit{Level 3} & 0.21623 & 0.21283 & 0.21850 & 0.22612 & 0.22791 \\
%     \midrule
%     Euclidean & \textit{Level 1} & 0.76111 & 0.83793 & 0.84664 & 0.87080 & 0.87210 \\
%     & \textit{Level 2} & 0.34202 & 0.36218 & 0.37751 & 0.48133 & 0.49511 \\
%     & \textit{Level 3} & 0.15349 & 0.17568 & 0.16543 & 0.22521 & 0.22767 \\
%     \bottomrule 
%   \end{tabular}
% \label{table:ldim-ablation-table}
% \end{table}

\subsection{Biologically-inspired synthetic dataset and the irregular variant}
Each 3D image of our biologically-inspired synthetic dataset consists of three levels of hierarchy. The first level of hierarchy (\textit{Level 1}) has a noisy background and an outer sphere of radius $r \sim \mathcal{N}(25, 1)$. Using a cell analogy, this represents the entire cell. The second level (\textit{Level 2}) consists of spheres (``vesicles'') and cuboids (``mitochondria''). Their sizes are randomly sampled with radius of $r \sim \mathcal{N}(8, 0.5)$ and with side length of $s \sim 2 \cdot \mathcal{N}(8, 0.5)$, respectively. In the third level (\textit{Level 3}) we introduce small spheres and cuboids (``protein aggregates'') in the vesicle spheres and mitochondria cuboids respectively. The \textit{Level 3} proteins have a radius of $r \sim \mathcal{N}(2, 0.2)$ and side length of $s \sim 2 \cdot \mathcal{N}(3, 0.15)$, respectively. The location of each object is sampled randomly, subject to the restriction that objects in Level $i + 1$ are entirely contained within an object in Level $i$. 

Each instance of a shape with a particular size is also given its own unique texture to mimic the different organelles of the cell. The color of each object is chosen randomly, according to a standard normal distribution. We also apply pink noise with magnitude $m=0.25$ to the volume as it is commonly seen in biological data. 

\begin{figure}
  \centering
  \includegraphics[width=0.8\linewidth]{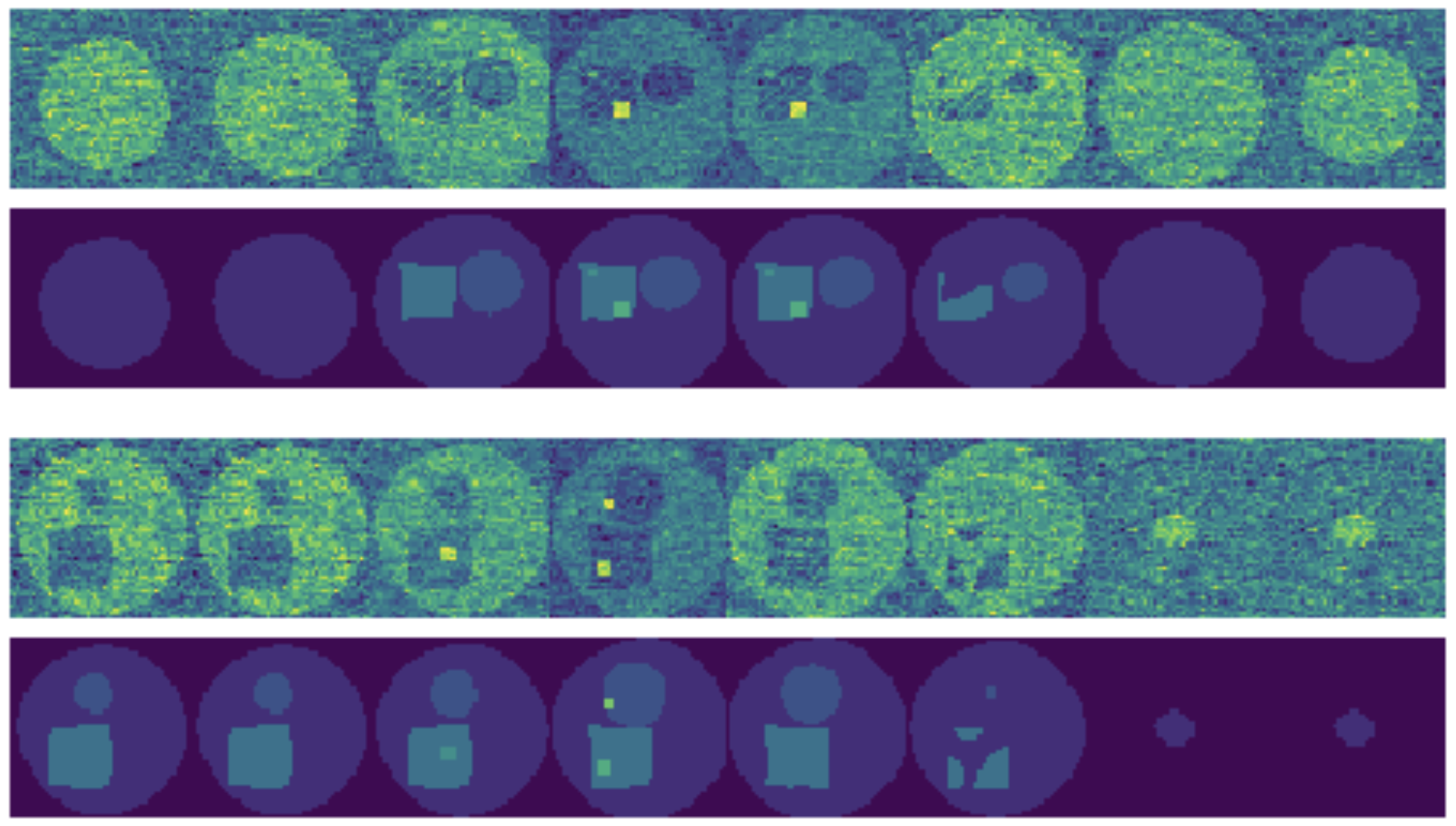}
  \caption{Sampled 2D slices from two examples of 3D volumes in our irregular biologically-inspired synthetic dataset, showing large variance in shapes across input. For each 3D volume example, the top row showcases the raw input data, and the bottom row showcases the ground truth segmentation.}
\label{fig:irregular}
\end{figure}

We generate an additional synthetic dataset with irregular shapes for evaluating datasets with large variance in characteristics across levels of hierarchy. This dataset was created through applying smooth noise to the boundaries of each shape. Specifically, we generate noise by first sampling random points in our voxel-grid and random values according to a Gaussian distribution, and interpolating to retrieve smooth noise. We then use this smooth noise function to perturb the points that fall within the interior of the three largest shapes. See an example of the dataset in Figure~\ref{fig:irregular}.

We demonstrate our method's performance in comparison to prior work on the aforementioned irregular dataset in Table~\ref{table:irregular_prior}, and an ablation study applied on the same irregular dataset in Table~\ref{table:irregular_ablation}.

We note that in Table~\ref{table:irregular_prior}, our proposed method outperforms prior work significantly on this irregular dataset, following our observations from our unperturbed synthetic dataset. We can see that while most methods show slight decrease in performance, our approach still shows state-of-the-art performance compared to prior unsupervised segmentation work across all hierarchical levels. 

For ablations on the irregular synthetic dataset in Table~\ref{table:irregular_ablation}, we find that our best models with hyperbolic latent space reliably outperform models with Euclidean latent space, as with our unperturbed synthetic dataset. Both Euclidean and hyperbolic base models have much lower performance on the irregular dataset compared to the unperturbed dataset, due to the challenges that the irregular dataset brings, for example, needing to recognize noisy instances of irregular shape as the same class. However, we demonstrate that the gyroplane convolutional layer and hierarchical triplet loss are both effective ways to improve performance on the base hyperbolic configuration. The inclusion of both of our contributions allows for significant performance gain across hierarchical levels, such that the results are comparable to that of the unperturbed dataset, even with a 23\% difference in \textit{Level 1} base hyperbolic performance.

\begin{table*}[htb!]
  \caption{Comparison with prior approaches on irregular synthetic dataset}
  \centering
  \begin{tabular}{llllll}
    \toprule
    & Dice \textit{Level 1} & Dice \textit{Level 2} & Dice \textit{Level 3} & Supervision type \\
    \midrule \citet{cciccek20163d} & 0.970 & 0.825 & 0.601 & 3D Semi-supervised \\ \citet{zhao2019data} & 0.978 & 0.641 & 0.333 & 3D Semi-supervised \\
    \midrule \citet{nalepa2020hyperspectral} & 0.559 & 0.259 & 0.138 & 3D Unsupervised \\ \citet{ji2019invariant} & 0.527 & 0.280 & 0.144 & 2D to 3D Unsupervised \\ \citet{moriya2018unsupervised} & 0.525 & 0.232 & 0.094 & 3D Unsupervised  \\
    \textbf{Ours} & \textbf{0.953} & \textbf{0.488} & \textbf{0.199} & 3D Unsupervised \\
    \bottomrule
  \end{tabular}
\label{table:irregular_prior}
\end{table*}

\begin{table*}[htb!]
  \caption{Ablation studies on irregular synthetic dataset}
  \centering
  \begin{tabular}{llllll}
    \toprule
    Latent Space & Configuration & Dice \textit{Level 1} & Dice \textit{Level 2} & Dice \textit{Level 3}  \\
    \midrule
    Euclidean & Base & 0.581 & 0.230 & 0.122 \\
     & Triplet & 0.823 & 0.392 & 0.175 \\
    \midrule
    Hyperbolic & Base & 0.607 & 0.284 & 0.158 \\
    & GyroConv & 0.812 & 0.401 & 0.197 \\
     & Triplet & 0.947 & \textbf{0.491} & 0.192 \\
     & GyroConv \& Triplet & \textbf{0.953} & 0.488 & \textbf{0.199}
     \\
    \bottomrule 
  \end{tabular}
\label{table:irregular_ablation}
\end{table*}

\subsection{BraTS ablations, error bars, and Hausdorff distance}

We conduct an ablation study on the BraTS dataset, with each of our added components with error bars over 4 independent runs. Results are shown in Table \ref{table:brats-ablation-table}. We can see that our best hyperbolic model outperforms our best Euclidean model significantly. The addition of the triplet loss improved both Euclidean and hyperbolic models, while the hyperbolic models see more performance gain. Our gyroplane convolutional layer also improves performance, while both of our additions jointly improve upon our Hyperbolic baseline, showing the benefit of these components to learning effective representations. 

\begin{table}[htb!]
  \caption{Ablation study for BraTS dataset. We report the mean and standard deviation of DICE scores for 4 independent runs.}
  \centering
  \begin{tabular}{lll}
    \toprule
    Latent Space & Configuration & Dice  \\
    \midrule
    Euclidean & Base & 0.388 $\pm$ 0.022 \\
     & Triplet & 0.517 $\pm$ 0.050 \\
    \midrule
    Hyperbolic & Base & 0.414 $\pm$ 0.017 \\
    & GyroConv & 0.539 $\pm$ 0.014 \\
     & Triplet & 0.610 $\pm$ 0.028 \\
     & GyroConv \& Triplet & \textbf{0.692} $\pm$ 0.009 \\
    \bottomrule 
  \end{tabular}
\label{table:brats-ablation-table}
\end{table}

We include the average and 95 percentile Hausdorff distance as complementary evaluation metrics on the BraTS dataset for comparison to prior unsupervised works in the main text. We describe the calculation below.

We use Hausdorff distance to evaluate the worst-case performance of our model. For two sets of points $A, B$, the directed Hausdorff distance from $A$ to $B$ is defined as 
\begin{align}\label{eqn:directed_hausdorff}
    h(A, B) = \max_{a \in A} \left\{ \min_{b \in B} \mathbf{d}(a, b) \right\}
\end{align}
where $\mathbf{d}$ is any distance function. We will take $\mathbf{d}$ to be the Euclidean distance. The Hausdorff distance is then defined to be
\begin{align}\label{eqn:hausdorff}
    H(A, B) = \max \left\{ h(A, B), h(B, A) \right\}
\end{align}
The official BraTS evaluation uses 95 percentile Hausdorff distance as measure of model robustness \citep{bakas2018identifying}.

%\begin{table}[hbt!]
%\caption{Comparison of our method against prior unsupervised work in Hausdorff distance. (Lower is better.)}
%\vskip 0.15in
%\begin{center}
%\begin{small}
%\begin{sc}
%\begin{tabular}{lll}
%    \toprule
%    & Average & 95\% \\
%    \midrule
%        \citet{moriya2018unsupervised} & 118.144 & 170.434 \\
%        \citet{ji2019invariant} & 96.865 & 114.400  \\
%        \cite{nalepa2020hyperspectral} & 87.704 & 110.803 \\
%    \textbf{Ours} & \textbf{77.940} & \textbf{97.641} \\
%    \bottomrule
%  \end{tabular}
%\end{sc}
%\end{small}
%\end{center}
%\vskip -0.1in
%\label{table:hausdorff}
%\end{table}

The BraTS dataset is licensed under Creative Commons Attribution.

\subsection{DICE score}
We use DICE score to quantitatively evaluate segmentation performance on all datasets. The DICE score is defined as the following:
\begin{align}
    DICE = \frac{2TP}{2TP + FN + FP}
\end{align}
where $TP$ is the number of true positives, $FN$ is the number of false negatives, and $FP$ is the number of false positives. For our synthetic dataset, we first assign predicted classes to ground truth labels using the Hungarian algorithm \cite{kuhn1955hungarian}, then evaluate using the average class DICE score. For the BraTS dataset \cite{menze2014multimodal, bakas2017advancing, bakas2018identifying}, we evaluate DICE of the whole tumor segmentation following official evaluation guidelines.

\subsection{Qualitative results in electron tomography}

\begin{figure*}[ht]
\vskip 0.2in
\begin{center}
\centerline{\includegraphics[width=14cm]{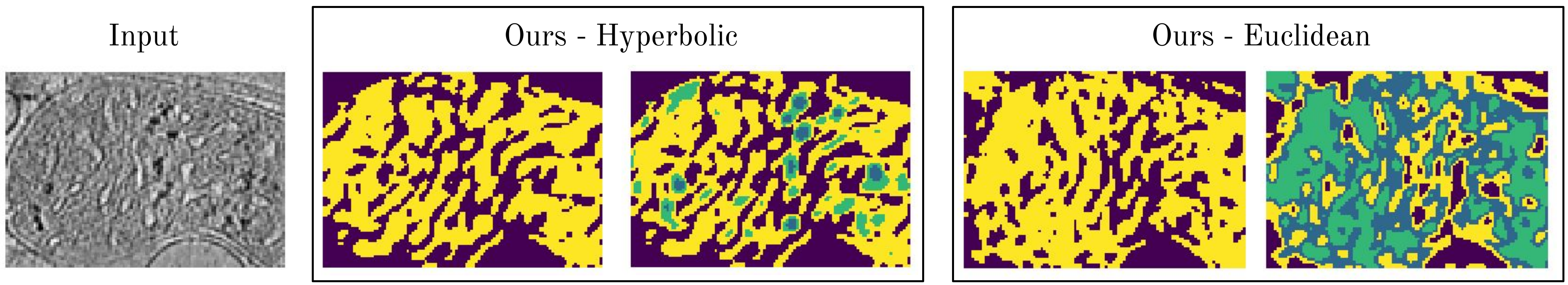}}
\caption{Leftmost image is a partial slice from a 3D cryo-ET image. The features of interest to be segmented are the dark densities with irregular shapes and sizes. The middle box shows segmentation from our best hyperbolic model, the rightmost box shows segmentation from our best Euclidean model. The segmentations in each box correspond to clustering using 2 vs. 4 classes.}
\label{fig:cryoem-figure}
\end{center}
\vskip -0.2in
\end{figure*}

\label{cryoem_section}

We show a real-world example where unsupervised segmentation of new biological organelles is important. Cryogenic electron tomography (cryo-ET) is a technique that images cells at cryogenic temperatures with a beam of electrons. The value of each voxel is the density at that location, and is created through reconstruction from tilt slices of $\pm 60$ degrees from electron tomography. 
Cryo-ET images are a rich source of biological data, capturing many unknown subcellular objects that we would like to identify and understand.

We train our model on three $512 \times 512 \times 250$ cryo-ET tomograms of cells collected from a research laboratory, and run inference on a fourth tomogram. Figure~\ref{fig:cryoem-figure} shows segmentations produced by our model on a mitochondria from the evaluation tomogram, using our proposed hyperbolic model vs. Euclidean model, and at a coarse and finer level of granularity. Unlike the Euclidean approach, our hyperbolic approach discovers a fine-grained class corresponding to small features on the mitochondria, which may be macromolecular aggregates. We can now investigate the discovered features for their biological identities and functions, leading to greater scientific understanding.

\subsection{Hyperparameters}
We use a single set of hyperparameters on all of our evaluation datasets, and these hyperparameters are not tuned on any of the evaluation datasets. In order to obtain a reasonable set of hyperparameters, we created a completely separate synthetic dataset on which we trained models and tuned hyperparameters. This synthetic dataset was created in a similar manner to our synthetic dataset; however, we designed it to have different and fewer objects, simpler nesting structure, no noise, and fewer textures. The application of this single set of hyperparameters to our evaluation datasets --- our synthetic dataset, the BraTS dataset, and the cryogenic electron tomography dataset, demonstrates the robustness of our approach. 

With the separate synthetic dataset that we used for choosing hyperparameters, we tuned over a range of values using its validation set. This includes weight of triplet loss in the range of $\beta = \{10^{-2}, 10^{-1}, 1, 10^{1}, 10^{2}, 10^{3}, 10^{4}, 10^{5}\}$, patch size for inference $p = \{5, 10, 15, 20, 40\}$, and number of epochs $e = \{3, 5, 8, 10, 12, 15\}$. We then used optimal hyperparameters $\beta = 10^3$, $p = 5$, and $e = 8$ for all experiments in our evaluation datasets. We used the Adam optimizer, w/ learning rate 1e-4, $\beta_1$= 0.9, $\beta_2$ = 0.999. Training time of our model is between 5 to 8 hrs on a Titan RTX GPU.

\subsection{Reproducibility of prior work}
Where available, we have used the authors' original code to generate the unsupervised baselines for the prior work comparisons. To sanity-check the code we used, we re-ran original experiments from the baseline paper. For \citep{ji2019invariant}, we re-ran their Potsdam-3 experiment for unsupervised 2D segmentation, and were able to reproduce the result from their paper to within approximately 1\%. For \citep{moriya2018unsupervised}, neither the original code nor the original dataset are publicly available, making reproducibility impossible to check, however we used the code base which their method was based on to implement their work. For \citep{nalepa2020hyperspectral}, the original code is unavailable as well, and we have adapted their method to our architecture in order to ensure a fair comparison.

\bibliographystyle{abbrvnat}
\bibliography{zappendix}

\section*{Checklist}

\begin{enumerate}

\item For all authors...
\begin{enumerate}
  \item Do the main claims made in the abstract and introduction accurately reflect the paper's contributions and scope?
    \answerYes{}
  \item Did you describe the limitations of your work?
    \answerYes{}
  \item Did you discuss any potential negative societal impacts of your work?
    \answerYes{See Appendix}
  \item Have you read the ethics review guidelines and ensured that your paper conforms to them?
    \answerYes{}
\end{enumerate}

\item If you are including theoretical results...
\begin{enumerate}
  \item Did you state the full set of assumptions of all theoretical results?
    \answerNA{}
	\item Did you include complete proofs of all theoretical results?
    \answerNA{}
\end{enumerate}

\item If you ran experiments...
\begin{enumerate}
  \item Did you include the code, data, and instructions needed to reproduce the main experimental results (either in the supplemental material or as a URL)?
    \answerNo{Instructions to reproduce results can be found in Implementation details in Section 4. The BraTS dataset is open source. We also plan to publicly release all code and synthetic datasets.}
  \item Did you specify all the training details (e.g., data splits, hyperparameters, how they were chosen)?
    \answerYes{See Section 4 and the Appendix.}
	\item Did you report error bars (e.g., with respect to the random seed after running experiments multiple times)?
    \answerYes{See Appendix.}
	\item Did you include the total amount of compute and the type of resources used (e.g., type of GPUs, internal cluster, or cloud provider)?
    \answerYes{See Appendix. All models are trained with 1 Titan RTX GPU.}
\end{enumerate}

\item If you are using existing assets (e.g., code, data, models) or curating/releasing new assets...
\begin{enumerate}
  \item If your work uses existing assets, did you cite the creators?
    \answerYes{}
  \item Did you mention the license of the assets?
    \answerYes{}
  \item Did you include any new assets either in the supplemental material or as a URL?
    \answerNo{We plan to publicly release both new synthetic datasets.}
  \item Did you discuss whether and how consent was obtained from people whose data you're using/curating?
    \answerNA{}
  \item Did you discuss whether the data you are using/curating contains personally identifiable information or offensive content?
    \answerNA{}
\end{enumerate}

\item If you used crowdsourcing or conducted research with human subjects...
\begin{enumerate}
  \item Did you include the full text of instructions given to participants and screenshots, if applicable?
    \answerNA{}
  \item Did you describe any potential participant risks, with links to Institutional Review Board (IRB) approvals, if applicable?
    \answerNA{}
  \item Did you include the estimated hourly wage paid to participants and the total amount spent on participant compensation?
    \answerNA{}
\end{enumerate}

\end{enumerate}